\RequirePackage{fix-cm}
\documentclass{svjour3}                     
\smartqed  
\usepackage{times}
\usepackage{soul}
\usepackage{url}
\usepackage[hidelinks]{hyperref}
\usepackage[utf8]{inputenc}
\usepackage[small]{caption}
\usepackage{graphicx}
\usepackage{amsmath}
\usepackage{booktabs}
\usepackage{color}
\usepackage{multirow}
\usepackage{amssymb}
\usepackage{enumerate}
\usepackage[table]{xcolor}
\usepackage[caption=false]{subfig}

\begin{document}

\title{Spatio-Temporal Adversarial Learning for Detecting Unseen Falls
}


\author{Shehroz S. Khan \and Jacob Nogas \and Alex Mihailidis}


\institute{
Shehroz S. Khan \at
              KITE -- Toronto Rehabilitation Institute, \\
              University Health Network, Canada. \\
              \email{shehroz.khan@uhn.ca}           
          \and
          Jacob Nogas, Alex Mihailidis \at
              University of Toronto, Canada.\\
              \email{jacob.nogas@mail.utoronto.ca, alex.mihailidis@utoronto.ca}
             \\
}

\date{Received: date / Accepted: date}

\maketitle

\begin{abstract}
Fall detection is an important problem from both the health and machine learning perspective. A fall can lead to severe injuries, long term impairments or even death in some cases. In terms of machine learning, it presents a severely class imbalance problem with very few or no training data for falls owing to the fact that falls occur rarely. In this paper, we take an alternate philosophy to detect falls in the absence of their training data, by training the classifier on only the normal activities (that are available in abundance) and identifying a fall as an anomaly. To realize such a classifier, we use an adversarial learning framework, which comprises of a spatio-temporal autoencoder for reconstructing input video frames and a spatio-temporal convolution network to discriminate them against original video frames. 3D convolutions are used to learn spatial and temporal features from the input video frames. The adversarial learning of the spatio-temporal autoencoder will enable reconstructing the normal activities of daily living efficiently; thus, rendering detecting unseen falls plausible within this framework. We tested the performance of the proposed framework on camera sensing modalities that may preserve an individual's privacy (fully or partially), such as thermal and depth camera. Our results on three publicly available datasets show that the proposed spatio-temporal adversarial framework performed better than other baseline frame based (or spatial) adversarial learning methods.
\end{abstract}


\section{Introduction}
Falls can cause severe injuries to people resulting in permanent or partial disability, huge health care costs and development of negative social and psychological problems \cite{smartrisk}. This constitutes a strong motivation to detect falls. However, a fall occurs rarely in comparison to normal activities of daily living (ADL) \cite{khan2017detecting}. Due to their rarity of occurrence, traditional supervised machine learning classifiers are difficult to use for this task \cite{khan2017review}. In many cases, there may be very few or no fall data available during training because collecting fall data is very challenging and can put people's life in danger \cite{khan2017review}. On the other hand, normal ADL performed by people are abundantly available and easier to collect. Therefore, we propose to detect falls in a one-class classification (OCC) framework \cite{khan2014one} that enables a classifier to learn only from the normal ADL and be able to detect an unseen fall during testing (as they may not be present during training).

Learning one-class classifiers from video sequences of normal ADL to detect falls as anomaly is a challenging task \cite{nogasfall2018}.  Previous research suggests that autoencoders can effectively learn ‘normal’ ADL from wearable and computer vision data and be able to detect abnormal variations, such as falls, based on the reconstruction error \cite{khan2017detecting}\cite{nogasfall2018}.  For detecting falls from videos, spatio-temporal autoencoders have shown to perform well in comparison to 2-D convolutional autoencoders and general deep autoencoders \cite{nogas2019deepfall}.
Another challenge in video based fall detection is to preserve the privacy of the person, which traditional RGB cameras cannot provide. Thus, detecting falls in videos without explicitly knowing a person's identity is important from the real world usability of such systems.

Convolutional neural networks (CNN), recurrent neural networks and spatio-temporal convolutional neural networks are commonly used to detect human activities \cite{zhao2017pooling}\cite{xu2018sequential}\cite{ji20133d} and anomalies \cite{sabokrou2018deep}\cite{zhou2019anomalynet}\cite{zhou2016spatial} in videos . The adversarial learning framework using different neural network models has also been used effectively to solve anomaly detection problems in images \cite{akcay2018ganomaly}\cite{schlegl2019f}\cite{schlegl2017unsupervised} and videos \cite{vu2019robust} \cite{ravanbakhsh2017abnormal}. 
The learning paradigm using generative adversarial networks (GAN) presents a unique opportunity to not only mimic normal behaviour through the generator but also to effectively discriminate it from anomalies \cite{schlegl2017unsupervised}. In the context of fall detection problem, adversarial learning will help in mimicking the normal ADL with high accuracy, which could result in detecting unseen falls with a higher degree of confidence. Therefore, in this paper, we extend the idea of training spatio-temporal autoencoder in an adversarial manner to validate their role in detecting unseen falls from (privacy protecting) videos. The proposed framework is different from the original formulation of GAN for anomaly detection, where images are generated from Gaussian noise \cite{schlegl2017unsupervised}.

In this paper, we design a new spatio-temporal adversarial learning framework, which consists of a spatio-temporal convolutional autoencoder (3DCAE) to reconstruct a sequence of normal ADL video frames and a spatio-temporal convolutional neural network (3DCNN) as a classifier to discriminate them from the original sequence of video frames. The spatio-temporal architecture of the adversarial framework consists of 3D convolutional layers that will extract both spatial and temporal features from the video frames  that will result in a robust system to learn normal ADL from the video sequences. After the training is completed, the 3DCAE would be able to reconstruct ADL sequences efficiently and the 3DCNN would be able to differentiate between real and reconstructed ADL sequences. 
During testing, when a video sequence containing fall frames is shown to this network, high reconstruction error and/or low probability of the discriminator will indicate an anomalous video sequence.
Therefore, this framework would be able to identify unseen falls with high accuracy. The reconstruction error of the 3DCAE  or the probability output of the 3DCNN or their combination can be used as an anomaly score to identify unseen falls during testing. 
We use two computer vision sensing modalities, thermal and depth cameras, to test the proposed framework. Both these sensing modalities can partially or fully obfuscate the facial identify of the person; thus, they are more promising to be used in a home-setting. 
We also implemented two spatial (or frame-based) variations of adversarial learning baselines with (i) a deep autoencoder to reconstruct input frames and a deep neural network as a discriminator, and (ii) a convolutional autoencoder to reconstruct input frames and a CNN as a discriminator (similar to the work of \cite{Sabokrou2018Adversarially}). 
The input to both of these methods is a frame from the video, whereas the input to our proposed method is a sequence of video frames.
Our results on three publicly available fall detection datasets captured using thermal and depth cameras show superior performance of the spatio-temporal adversarial learning framework in detecting unseen falls in comparison to these spatial adversarial approaches.

The paper is organized as follows. In Section \ref{sec:litrev}, we present literature review on using adversarial techniques for anomaly detection in images and videos. In Section \ref{sec:adversarial}, we introduce the proposed spatio-temporal adversarial learning framework. Section \ref{sec:unseenfalls} presents various anomaly scores to detect unseen falls. The experiments and results are described in Section \ref{sec:results}, followed by conclusions and pointers to future research in Section \ref{sec:conclusions}.

\section{Related Work}
\label{sec:litrev}

In this paper, we detect falls in an OCC framework. To the best of our knowledge, fall detection has not been addressed using an adversarial learning framework; therefore, we present related literature review on techniques that use adversarial learning of autoencoders (or their variants) for general anomaly detection in images and videos. 

\subsection{Adversarial Anomaly Detection in Images}

One of the earliest work to detect anomalies using adversarial framework is presented by Schlegl et al. \cite{schlegl2017unsupervised} to find anomalies in imaging data as candidates for markers, called as \textit{AnoGan}. The generator of their GAN is equivalent to a multi-layered convolutional decoder that samples input from uniformly distributed noise. The discriminator is a standard CNN that maps 2D images to a single value that can be interpreted as a probability whether the input to it is a real image or is produced by the generator. They use the combination of residual and discrimination losses as an anomaly score, such that a large score means an anomalous image.  They extended their approach by presenting a faster anomaly detection algorithm (\textit{f-AnoGAN} \cite{schlegl2019f}) that used improved WGAN architecture and speed up  mapping of input images to the latent space. Beggel et al. \cite{Beggel2019Robust} considered identifying anomalies in images when the training set is contaminated with a small fraction of outliers. They trained an adversarial autoencoder that imposed a prior distribution on the latent representation by placing anomalies in the low likelihood-region. This architecture helped in identifying potential anomalies and robust detection in the presence of outliers during training. Pidhorskyi \cite{NIPS2018_7915} presented a probabilistic approach to adversarial training of autoencoders for anomaly detection by estimating the likelihood of a sample being generated by the inlier distribution. This was achieved by linearizing the parameterized manifold capturing the underlying structure of the inlier distribution and improved autoencoder training. Their results on several publicly available image datasets show improved results.

Eide \cite{eide2018applying} applied generative adversarial learning to find anomalies in hyper-spectral remote sensing images. Their generator is based on ResNet, which maps low-dimensional input to a higher dimension image; thus, works as a convolutional decoder. The discriminator has a similar design but works in the opposite direction. They modify the reconstruction cost of the generator by adding a term for the norm of generated input. The modified reconstruction cost penalizes reconstructions from unlikely inputs more heavily. However, adding this term is not found to be helpful as the generator is unable to reconstruct anomalies even without any penalty term. 
Yarlagadda et al. \cite{yarlagadda2018satellite} present the use of adversarial autoencoder learning for satellite image forgery detection and localization. The generator in their structure is an autoencoder and the discriminator is a CNN. The adversarial trained autoencoder encodes the image patches into low dimensional features, which are then used to train a one-class SVM to detect forged patches.
Lawson et al. \cite{lawson2017finding} present the use of adversarial trained deep convolutional autoencoder for finding anomalies in autonomous robot patrol view. Their method first learns the model for normal scene from the autoencoder based generator and then use the features learned to find anomalies in the environment. More specifically, they compare the difference between the bottleneck features extracted with real images and reconstructed images and use it as a measure for finding anomalies.

\subsection{Advesarial Anomaly Detection in Videos}

Sabokrou et al. \cite{Sabokrou2018Adversarially} present an end-to-end OCC method that uses adversarial learning. The generator of their network is a convolutional autoencoder, which reconstructs the input with added noise. The discriminator is a typical CNN that takes reconstructed and real input and gives a likelihood estimate of the target score. After the adversarial training, the discriminator can be used to detect anomalies. They also show that applying discriminator on the reconstructed images can provide better separation; hence, better performance. Their results on MNIST, Caltech-256 and UCSD Ped2 datasets show the viability of learning one-class classifiers in an adversarial manner.
Lee et al. \cite{lee2018stan} present a spatio-temporal adversarial learning framework for anomaly detection in videos. Their framework consists of a spatio-temporal generator and discriminator. The network operates on a sequence of $N+1$ video frames. The generator takes as input the first and last $\frac{N}{2}$ frames and then generates the missing $\frac{N}{2}+1^{th}$ frame. This middle frame is generated by a bi-directional convolutional LSTM network. The discriminator consists of a 3DCNN that takes a sequence of $N+1$ frames as input, which has one generated frame and rest are original frames. The discriminator then tries to recognize this sequence as fake, while the generator must improve to generate the middle frame in order to fool the discriminator. A potential issue with such an approach is that the discriminator is given a very difficult task to only detect one frame in a sequence; and conversely the generator is given an easy task. 
Vu et al. \cite{vu2019robust} presented a multi-level representation of intensity and motion in videos to identify anomalies. Their framework consisted of a de-noising autoencoder, conditional generative adversarial network and anomalous region detector at each representation level. Besides showing improved results on UCSD Ped 1, UCSD Ped 2 and Avenue video anomaly datasets, their model was able to detect mislabeled anomalies in UCSD Ped 1 dataset. Li and Chang \cite{li2019video} presented an approach to train Multivariate Gaussian Fully Convolution Adversarial Autoencoder to map the latent space representations of normal samples. A deep CNN was employed for the encoder of the deep network, then an energy based method is applied to obtain anomaly score. The appearance and motion representations were combined to obtain robust anomaly detection results on three public datasets. Liu et al. \cite{liu2018future} used the difference between future frame prediction and ground truth as a factor to detect anomalies in videos. Their objective function combines different losses, including appearance (intensity loss and gradient loss), motion (optical flow loss) and adversarial loss. They adopted a U-net \cite{ronneberger2015u} as a generator and a Markovian discriminator \cite{isola2017image} in their framework. Li et al. \cite{li2019spatio} presented a U-net based frame prediction method using normal events in videos and detecting abnormality using prediction error. They considered different types of losses in their objective function that includes intensity, gradient, motion, RGB gradient and a mean square error loss during adversarial training. Nguyen and Meunier \cite{nguyen2019anomaly} designed a video anomaly detection framework that  combines a Convolution Autoencoder and a U-Net that is integrated with an Inception module leading to a patch based frame level anomaly score. They trained this network using distance based loss, optical flow loss and adversarial loss. 
Ravanbaksh et al.\cite{ravanbakhsh2017abnormal} present the use of adversarial learning for anomaly detection in crowded scenes. They train two conditional U-nets \cite{isola2017image}; one each for generating  optical flow from frames and the other generating frames from optical flow using image and noise vector as inputs. The conditional discriminator takes either of the generated images and compares against the real image to produce a probability that both of its input images come from the real data. However, this method may not work well with occluded scenes and it may be difficult to estimate the optical flow map. 
Tang et al. \cite{tang2020integrating} combined the future frame prediction and reconstruction error of two U-nets connected in a pipeline, combined with a pixel-level discriminator to detect anomalies in videos. They used intensity, gradient and temporal image difference losses and trained the network in an adversarial manner. They showed better results on several public datasets in comparison to the baseline presented by Liu et al. \cite{liu2018future}. 
Zhou et al. \cite{zhou2019attention} used attention based loss in an adversarial learning setting to alleviate the foreground-background imbalance problem in anomaly detection in videos. They considered U-net based  generator and a patch discriminator. Their results on public video datasets showed improvement in comparison to the baseline of Liu et al. \cite{liu2018future}. Some other variants of 3D GANs are also proposed for other applications. Wang et al. \cite{wang2017shape} combine 3D GAN with Recurrent Convolutional Networks for Shape Inpainting, and Zhang et al. \cite{zhang2019adversarial} present a 3D GAN for video deblurring.

The spatio-temporal adversarial learning method to detect unseen falls presented in this paper extends the work of Sabokrou et al. \cite{Sabokrou2018Adversarially} from single image to a sequence of images (video) by learning spatio-temporal features. The proposed frame also differs from the work of Lee et al. \cite{lee2018stan}  in that it uses a 3DCAE instead of the bi-directional convolutional LSTM or 2D CAE. The work of Nogas et al. \cite{nogas2019deepfall} suggests that training LSTM based autoencoders can be slower in comparison to 3DCAE.
Our 3DCAE reconstructs the whole sequence of frames  given an input sequence of frames instead of producing only one frame and is fed to the 3DCNN discriminator. This way the discriminator is presented with a fully reconstructed sequence of frame, rather than one frame in a sequence to decide if its real or reconstructed. In the next section, we describe the various components of the proposed spatio-temporal adversarial framework for detecting unseen falls.

\section{Spatio-Temporal Adversarial Learning}
\label{sec:adversarial}
The proposed spatio-temporal adversarial learning framework for identifying unseen falls consists of: (i) training a 3DCAE to reconstruct a sequence of normal ADL video frames and, (ii) a 3DCNN to discriminate the reconstructed sequences with the original sequences of video frames of normal ADL. Both of these components perform 3D Convolution operations. A 3D convolutional layer is defined as follows: the value $\boldsymbol{v}$ at position $(x,y,z)$ of the $j^{th}$ feature map in the $i^{th}$ 3D convolution layer, with bias $b_{ij}$, is given by the equation \cite{ji20133d}

\begin{equation}\label{eq:2}
    \boldsymbol{v}_{ij}^{xyz} = f(\sum_m \sum_{p=0}^{P_i-1} \sum_{q=0}^{Q_i-1} \sum_{s=0}^{S_i-1}  \boldsymbol{w}_{ijm}^{pqs}v_{(i-1)m}^{(x+p)(y+q)(z+s)} + b_{ij})
\end{equation}
where $P_i$, $Q_i$, $S_i$ are the vertical (spatial), horizontal (spatial), and temporal extent of the filter cube $\boldsymbol{w}_i$ in the $i^{th}$ layer. The set of feature maps from the $(i-1)^{th}$ layer are indexed by $m$, and $w_{ijm}^{pqs}$ is the value of the filter cube at position $pqs$ connected to the $m^{th}$ feature map in the previous layer. Multiple filter cubes will output multiple feature maps. 
Next, we describe the 3DCAE and 3DCNN that are used in the proposed adversarial learning framework to detect unseen falls.

\subsection{3DCAE} 
The spatio-temporal autoencoder used in this paper, 3DCAE, is derived from the works of Nogas et al. \cite{nogas2019deepfall}. The specification of the convolution filters, number of layers and depth of the network have been reported to work well for the fall detection problem from videos. We used the same baseline and have not added confounding parameters to make the model complex. The input to 3DCAE, $\boldsymbol{I}$, comprises of a continuous sequence of $t=1,\ldots,T$ frames, called a window. These windows of length $T=8$ are generated by applying a temporal sliding window to input video frames, with padding (or not) and stride (the amount of frames shifted from one window to the next).
The input $\boldsymbol{I}$ is encoded by a sequence of 3D convolution layers. The first 3D convolution layer uses 3D convolutions with stride of $1 \times 2 \times 2$, and padding, and the rest use stride of $2 \times 2 \times 2$, and padding. This means that each dimension (temporal depth, height, and width) is reduced by a factor of $2$ with every 3D convolution layer except the first, which reduces only the spatial dimension, thus allowing for a deeper architecture without collapsing the temporal dimension completely. Decoding operates as encoding but in reverse, using 3D deconvolution layers. The final deconvolution layer combines feature maps into the decoded reconstruction. This final layer uses a stride of $1 \times 1 \times 1$ and padding.
For hidden layers, the activation function $f$ is set to ReLU.  We use $P_i = Q_i = 3$, and $S_i = 5$, for all convolutional and deconvolutional layers, as these values were found to produce the best results across all the datasets. Table \ref{tab:config3D} shows the configuration of the 3DCAE used in our spatio-temporal adversarial framework. The output of the 3DCAE (reconstructed video sequence of size $T$) is fed to the 3D discriminator along with the actual input video sequence of size $T$. 
Batch normalization is used in all the layers of the 3DCAE except for the final layer. 

\begin{table}[ht]
\centering
\begin{tabular}{|c|c|} \hline
\textbf{Input} &  (8, 64, 64, 1)  \\ \hline\hline
\multirow{4}{*}{\textbf{Encoder}}& 3D Convolution - (8, 64, 64, 16) \\ \cline{2-2}
& 3D Convolution - (8, 32, 32, 8)  \\ \cline{2-2}
& 3D Convolution - (4, 16, 16, 8) \\ \cline{2-2}
& 3D Convolution - (2, 8, 8, 8)  \\ \hline\hline
\multirow{4}{*}{\textbf{Decoder}}& 3D Deconvolution - (4, 16, 16, 8) \\ \cline{2-2}
& 3D Deconvolution - (8, 32, 32, 8) \\ \cline{2-2}
& 3D Deconvolution - (8, 64, 64, 16) \\ \cline{2-2}
& 3D Convolution - (8, 64, 64, 1)  \\ \hline
\end{tabular}
\caption{Configuration of the 3D Generator. The values inside the parenthesis for fully connected layers are the number of neurons.}
\label{tab:config3D}
\end{table}

\subsection{3D Discriminator} 
The discriminator in our setting is a 3DCNN, whose architecture is kept the same as the encoding configuration of the 3DCAE followed by a fully connected layer of one neuron at the end with a sigmoid function to output a probability of whether a sequence of frames is original or reconstructed. Batch normalization is used in all the layers of the 3D discriminator except for  the input layer. LeakyRelu activation is set in all hidden layers, with negative slope coefficient set to $0.2$.

It is to be noted that during the training phase, only the video sequences of normal ADL are presented to the 3DCAE and 3DCNN, whereas during testing phase video sequences may contain both normal ADL and fall frames.

\subsection{Adversarial Learning}
\begin{figure}[!ht]
    \centering
    \includegraphics[width=\textwidth]{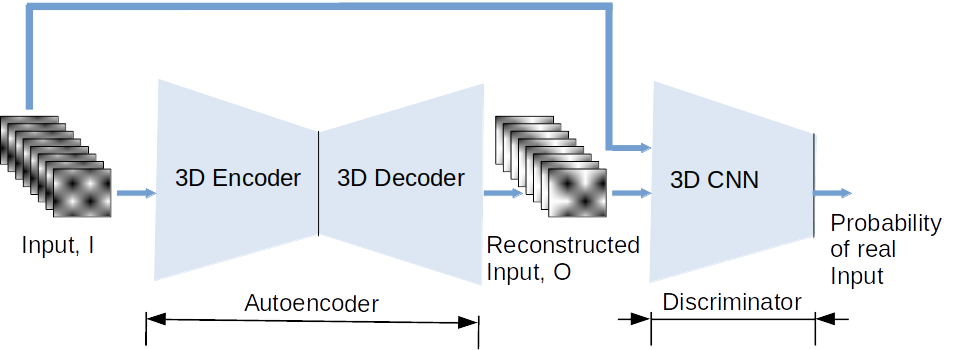}
    \caption{Block diagram of spatio-temporal adversarial framework to detect unseen falls.}
    \label{fig:3DGAN}
\end{figure}

As discussed previously, the proposed adversarial framework consists of two components; a 3DCAE as a generator and a 3DCNN as a discriminator. Figure \ref{fig:3DGAN} shows the setup of the overall adversarial framework, where the autoencoder and discriminator are trained in an adversarial setting. The 3DCAE (represented as $\mathcal{R}$) takes the input sequence ($\boldsymbol{I}$) of window size $T$ of normal ADL, and reconstructs the sequence, $\boldsymbol{O}$, which is then fed to fool 3DCNN (represented as $\mathcal{D}$) that it is an original input and not the reconstructed sequence. However, $\mathcal{D}$ will have access to the original input sequence ($\boldsymbol{I}$) and may easily identify the reconstructed sequence as not the original input sequence. Then, the two components play an adversarial game, which after completion of training should enable $\mathcal{R}$ to reconstruct input video sequences with minimum reconstruction error to successfully fool $\mathcal{D}$. This means that $\mathcal{R}$ should be able to reconstruct output sequence very similar to the input sequence. In other words, the spatio-temporal autoencoder would have learned the concept of normal ADL after successful completion of the training. This further implies that any sequence with anomaly (e.g. fall) would be reconstructed with high reconstruction error. At the same time, the discriminator would have become an expert to identify between the badly reconstructed sequences and the input sequences.

In our setting, $\mathcal{R}$ maps $\boldsymbol{I}$ to $\boldsymbol{O}$ using the distribution of the target class $p$, i.e.

\begin{equation}
\boldsymbol{O} = \boldsymbol{I} \sim \boldsymbol{p}
\end{equation} 

However, $\mathcal{D}$ has access to input samples and is exposed to $p$. Therefore, $\mathcal{D}$ can explicitly decide if $\mathcal{R(O)}$ comes from $p$ or not. The objective function to jointly learn $\mathcal{R}$ and $\mathcal{D}$ can be written as:

\begin{equation}
    \min_\mathcal{R} \max_\mathcal{D} \quad (\mathbb{E}_{I \sim p} [log(\mathcal{D}(I))]) +
    \mathbb{E}_{O \sim p} [log(1 - \mathcal{D}(R(O)))])
\end{equation}

To train the model, we need to calculate the (i) loss due to the 3DCAE ($\mathcal{L_R}$), and (ii) loss due to both 3DCAE and the 3DCNN ($\mathcal{L_{R+D}}$). The 3DCAE loss is simply the reconstruction error between the $j^{th}$ frame of $I_i$ and $O_i$, and can be written as

\begin{equation}
    \mathcal{L_R}= \Vert \boldsymbol{I}_{i,j} - \boldsymbol{O}_{i,j} \Vert_2^2
    \label{eq:reconerror}
\end{equation}

Thus, the total loss function to minimize can be written as:

\begin{equation}
    \mathcal{L} = \mathcal{L}_\mathcal{R} + \lambda \mathcal{L}_{\mathcal{R}+\mathcal{D}}
\end{equation}

where $\lambda$ is a positive number that controls the relative importance of both the loss terms.

For comparison purposes, we implement two other variants of autoencoders to detect unseen falls that are trained as per the proposed  adversarial framework. The first variant uses a deep autoencoder as a generator and a multi-layer feed forward network as the discriminator, we call it as DAE-AN. The configuration of the discriminator is the same as the encoder of the deep autoencoder. This method will learn global features from the video sequences to successfully reconstruct ADL. The second variant uses a convolutional autoencoder (CAE) as a generator and a convolutional feed-forward network as a discriminator, we call it as CAE-AN. The configuration of the discriminator, in this case, is the same as the encoder of the CAE (this framework is analogous to the work of Sabokrou et al. \cite{Sabokrou2018Adversarially}). This method will learn localized spatial features. The structure of the encoder and decoder for DAE-AN and CAE-AN are shown in Tables \ref{tab:configDAE} and \ref{tab:configCAE}.
It is to be noted that the input to the DAE-AN and CAE-AN is a frame from the video, whereas the input to the proposed spatio-temporal adversarial learning method is a window consisting of a sequence of $T$ frames, as shown in Figure \ref{fig:3DGAN}. Therefore, the proposed method will learn both spatial and temporal features when the training is successfully completed.

\begin{table}[!ht]
\centering
\begin{tabular}{|c|c|} \hline
\textbf{Input} &  (64, 64, 1)  \\ \hline\hline
\multirow{4}{*}{\textbf{Encoder}}& Fully Connected - (4096) \\ \cline{2-2}
& Fully Connected - (1500)  \\ \cline{2-2}
& Fully Connected - (1000) \\ \cline{2-2}
& Fully Connected - (500)  \\ \hline\hline
\multirow{4}{*}{\textbf{Decoder}}& Fully Connected - (1000) \\ \cline{2-2}
& Fully Connected - (1500) \\ \cline{2-2}
& Fully Connected - (4096) \\ \cline{2-2}
& Fully Connected - (64, 64, 1)  \\ \hline
\end{tabular}
\caption{Configuration of the DAE-AN . The values inside the parenthesis for fully connected layers are the number of neurons.}
\label{tab:configDAE}
\end{table}

\begin{table}[!ht]
\centering
\begin{tabular}{|c|c|} \hline
\textbf{Input} &  (64, 64, 1)  \\ \hline\hline
\multirow{4}{*}{\textbf{Encoder}}& 2D Convolution - (64, 64, 16) \\ \cline{2-2}
& 2D Convolution - (32, 32, 16)  \\ \cline{2-2}
& 2D Convolution - (16, 16, 8) \\ \cline{2-2}
& 2D Convolution - (8, 8, 8)  \\ \hline\hline
\multirow{4}{*}{\textbf{Decoder}}& 2D Deconvolution - (16, 16, 8) \\ \cline{2-2}
& 2D Deconvolution - (32, 32, 8) \\ \cline{2-2}
& 2D Deconvolution - (64, 64, 16) \\ \cline{2-2}
& 2D Deconvolution - (64, 64, 1)  \\ \hline
\end{tabular}
\caption{Configuration of the CAE-AN. The values inside the parenthesis are the size of the convolution filters.}
\label{tab:configCAE}
\end{table}

\begin{figure}[!ht]
    \centering
    \hspace{25mm}
    \includegraphics[scale=0.45]{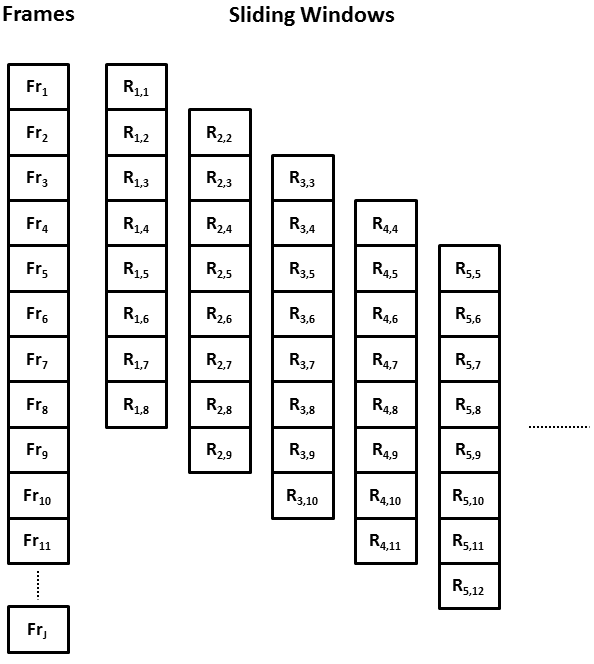}
    \caption{Temporal sliding window showing reconstruction error ($R_{i,j}$) per frame ($Fr_j$) with $T=8$ .}
    \label{fig:windows}
\end{figure}

\section{Detecting Unseen Falls}
\label{sec:unseenfalls}
The spatio-temporal framework is trained in an adversarial manner on only normal ADL and an unseen fall is detected as an anomaly during testing. 
The method to detect unseen falls is shown in Figure \ref{fig:windows} (derived from \cite{nogas2019deepfall}). All the frames in the video, $Fr_i$, are broken down into windows of frames of length, $T=8$, with stride=$1$. 
For the $i^{th}$ window $\boldsymbol{I}_{i}$, the 3DCAE outputs a reconstruction of this window, $\boldsymbol{O}_{i}$. The reconstruction error ($R_{i,j}$) between the $j^{th}$ frame of $I_i$ and $O_i$ can be calculated as (similar to Equation \ref{eq:reconerror})

\begin{equation}
    R_{i,j}= \Vert \boldsymbol{I}_{i,j} - \boldsymbol{O}_{i,j} \Vert_2^2
\end{equation}

There are two ways to detect unseen falls, (i) at the frame level, or (ii) at the window level, which are described next. 

\paragraph{Frame Level Anomaly}: 
In the frame level anomaly method, the reconstruction error ($R_{i,j}$) (obtained from the 3DCAE) is computed for every frame $j$ across different windows. The average ($C_{\mu}^{j}$) and standard deviation ($C_{\sigma}^{j}$) of a frame $j$ across different windows are used as an anomaly score as follows \cite{nogas2019deepfall}:

\begin{equation}
\begin{aligned}
C_{\mu}^{j} &= \begin{cases}\frac{1}{j} \sum_{i=1}^j R_{i,j}  & j<T\\
\frac{1}{T} \sum_{i=1}^T R_{i,j} &  j \geq T
\end{cases}\\
 C_{\sigma}^{j} &= \begin{cases}\sqrt{\frac{1}{j} \sum_{i=1}^j (R_{i,j}-C_{\mu}^{j})}& j<T\\
 \sqrt{\frac{1}{T} \sum_{i=1}^T (R_{i,j}-C_{\mu}^{j})} & j \geq T
 \end{cases}
 \end{aligned}
\end{equation}

$C_{\mu}^{j}$ and $C_{\sigma}^{j}$ give an anomaly score per-frame, while incorporating information from the past and future frames. A large value of $C_{\mu}^{j}$ or $C_{\sigma}^{j}$ means that the $j^{th}$ frame, when appearing at different positions in subsequent windows, is reconstructed with a high average error or with highly variable error; thus, indicating the occurrence of a fall. As this method calculates anomaly at the frame level, it is directly comparable with DAE-AN and CAE-AN.
For DAE-AN and CAE-AN, the reconstruction error of an input frame is used as an anomaly score.

\paragraph{Window Level Anomaly}: In the window level anomaly method, the score for the entire window of $T$ frames is calculated. For an input $x$ comprising of $T$ frames, this score, can be either of the following:

\begin{enumerate}[(i)]
    \item Reconstruction error of the 3DCAE, $R(x_{i,j})$. For a particular window $i$, the mean of reconstruction error of all the $T$ frames ($W_{\mu}^{i}$) and their standard deviation ($W_{\sigma}^{i}$) are used as an anomaly score, as follows:

        \begin{equation}
            W_{\mu}^{i} = \frac{1}{T} \sum_{j=i}^{T+i-1} R_{i,j} ,\quad
            W_{\sigma}^{i} = \sqrt{\frac{1}{T} \sum_{j=i}^{T+i-1} (R_{i,j}-W_{\mu}^{i})}
        \end{equation} 

    \item Probability score of the discriminator 3DCNN, $\mathcal{D}(x)$,
    \item Probability score of the discriminator on the reconstructed input, $\mathcal{D}(\mathcal{R}(x))$ \cite{Sabokrou2018Adversarially},
    
    Combination of both autoencoder and discriminator scores, i.e.
    \item $\mathcal{D}(x) + \lambda \mathcal{R}(x)$, and
    \label{score:dx}
    \item $\mathcal{D}(\mathcal{R}(x)) + \lambda \mathcal{R}(x)$
    \label{score:drx}
\end{enumerate}

The anomaly scores (\ref{score:dx}) and (\ref{score:drx}) will have two versions each based on the mean and standard deviation of the reconstruction error, represented as $W_\mu - \mathcal{D}(x)+ \lambda \mathcal{R}(x)$ and $W_\sigma - \mathcal{D}(x)+ \lambda \mathcal{R}(x)$, and $W_\mu - \mathcal{D}(\mathcal{R}(x))+ \lambda \mathcal{R}(x)$ and $W_\sigma - \mathcal{D}(\mathcal{R}(x))+ \lambda \mathcal{R}(x)$. The $-$ sign should not be confused with the minus sign; it only shows that this particular scores is derived from the mean or standard deviation of the reconstruction error.

The number of fall frames present in a window ($\alpha$), s.t. the ground truth label of the entire window is a fall is a hyperparameter of the method and will influence the detection of anomalies. Giving a window the ground truth as a fall with low value of $\alpha$ may result in high false alarm rate.  Whereas deciding a window as a fall with high value of $\alpha$ may miss some falls. In the experiments, we varied the value of $\alpha$ from $1$ to $T$ to understand the impact of choosing its appropriate value.

\section{Experiments and Results}
\label{sec:results}

\subsection{Datasets}
We use the following three datasets to test the proposed spatio-temporal adversarial framework to detect unseen falls.
All of these datasets contain videos captured through thermal or depth cameras. Therefore, these datasets are capable of partially or fully obfuscating the identity of the person in the video.

\begin{enumerate}
    \item \textit{Thermal Dataset}:
    The Thermal dataset \cite{Thermal} contains $9$ videos with normal ADL and $35$ videos containing falls and other normal activities. These videos are captured using a FLIR ONE thermal camera mounted to an Android phone with a spatial resolution of $640 \times 480$. The videos have a frame rate of either 25 fps or 15 fps, which was obtained by observing the properties of each video. The thermal camera can protect the privacy/identity of the individual and can capture images during night conditions as well. To create sequence of windows to be given as input to the proposed spatio-temporal adversarial framework, sliding window ($T=8$) is performed on all video frames, resulting in $22,116$ frames from $9$ ADL videos. A sample of normal ADL and fall activities from the thermal dataset is shown in Figure \ref{fig:ThermalSamples}.
    
\begin{figure}[!ht]
\centering
\captionsetup[subfigure]{width=80pt, justification=centering}%
  \begin{subfloat}[]
    \centering
    \includegraphics[scale=0.15]{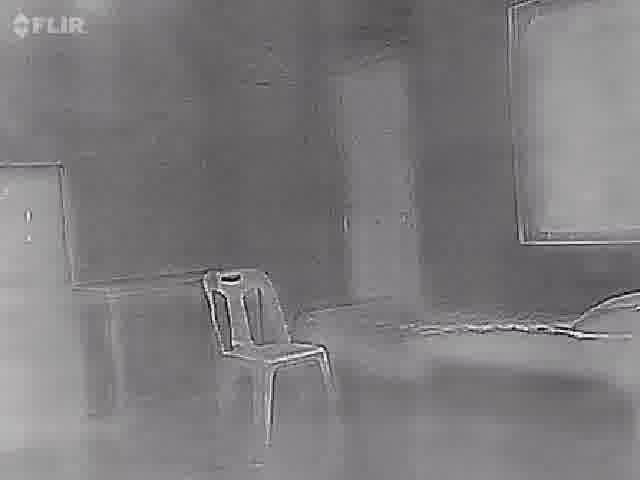}
    \label{fig:ThermalSample1}
  \end{subfloat}
  \begin{subfloat}[]
    \centering
    \includegraphics[scale=0.15]{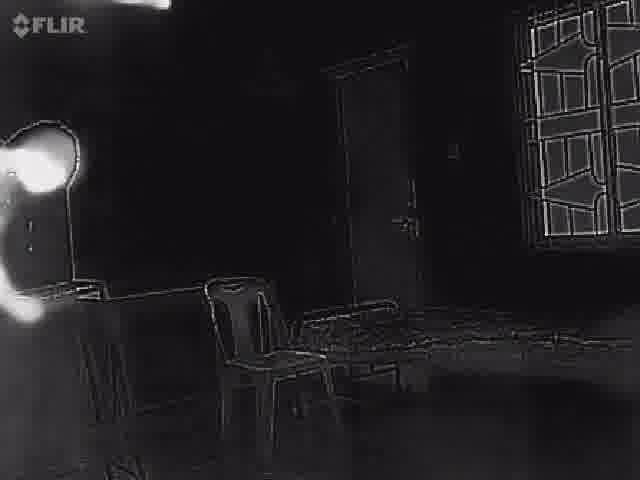}
    \label{fig:ThermalSample2}
  \end{subfloat}
    \begin{subfloat}[]
    \centering
    \includegraphics[scale=0.15]{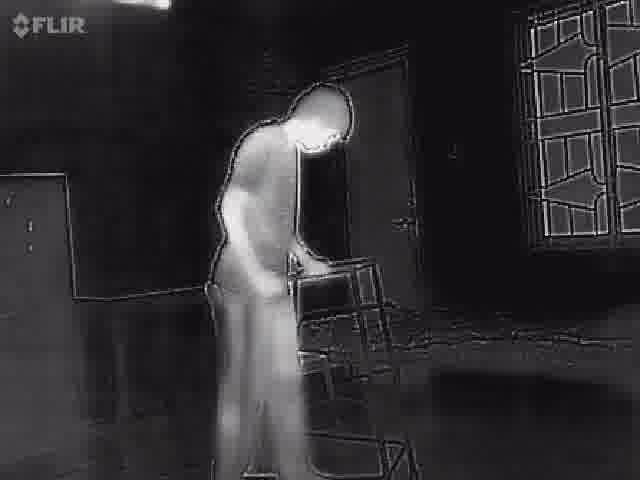}
    \label{fig:ThermalSamples3}
  \end{subfloat}
  \vspace{5mm}
      \begin{subfloat}[]
    \centering
    \includegraphics[scale=0.15]{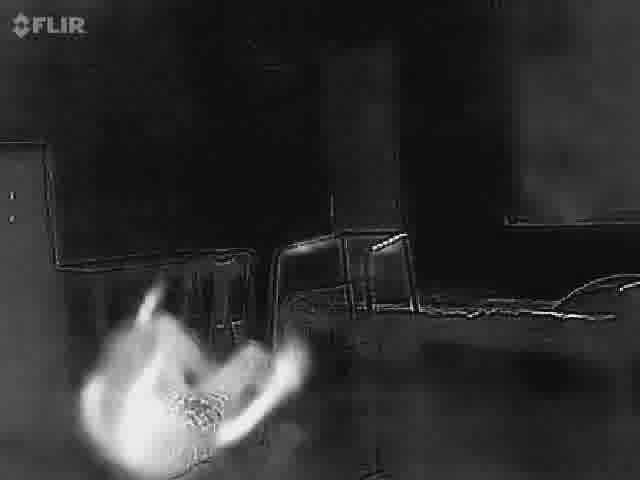}
    \label{fig:ThermalSamples4}
  \end{subfloat}
  \begin{subfloat}[]
    \centering
    \includegraphics[scale=0.15]{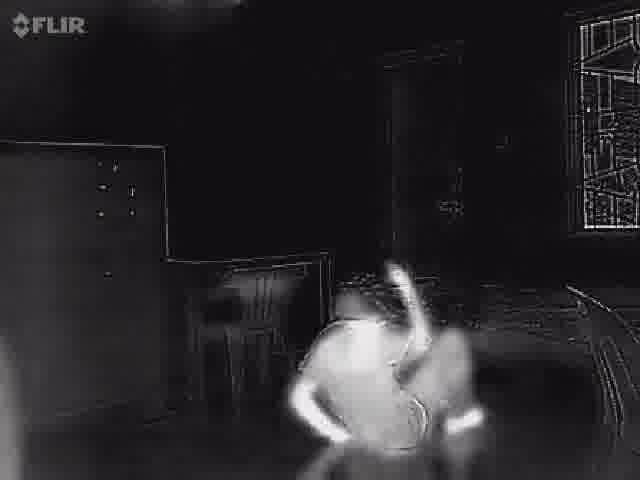}
    \label{fig:ThermalSamples5}
  \end{subfloat}
  \begin{subfloat}[]
    \centering
    \includegraphics[scale=0.15]{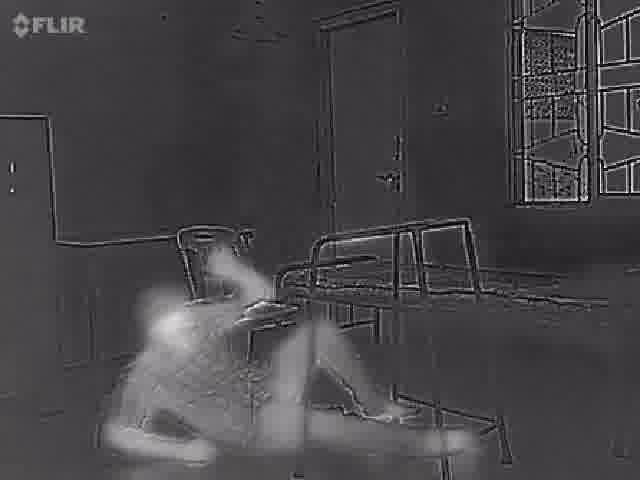}
    \label{fig:ThermalSamples6}
  \end{subfloat} 
  \centering
  \caption{Thermal Dataset -- ADL frames (a) Empty Scene, (b) Person Entering the Scene, (c) Person in the Scene,  and Fall Frames (d), (e), and (f).}
  \label{fig:ThermalSamples}
\end{figure}
 
\item \textit{UR Dataset}:
    The UR dataset \cite{UR} contains $40$ videos of person doing normal ADL (such as walking, sitting down, crouching down, and lying down in bed.) and $30$ videos with a fall in them. Two types of falls were performed by five persons from standing and sitting on the chair. These videos are captured at $30$ fps using a Kinect depth sensor, which obfuscate the identity of the person. The depth map is stored in VGA resolution ($640 \times 480$). The UR dataset has many missing pixel regions, called `holes', which were filled using a method based on depth colorization \cite{Silberman:ECCV12}. The new version of this dataset obtained after filling the holes is called as UR-filled in this paper. 
    After applying the sliding window ($T=8$), $8,661$ windows of contiguous ADL frames were obtained for training the spatio-temporal adversarial framework. A sample of normal ADL and falls from UR and UR-filled dataset is shown in Figures \ref{fig:URHoles} and \ref{fig:URHolesFilled}.
    
\begin{figure}[!ht]
\centering
\captionsetup[subfigure]{width=80pt}%
  \begin{subfloat}[]
    \centering
    \includegraphics[scale=0.1]{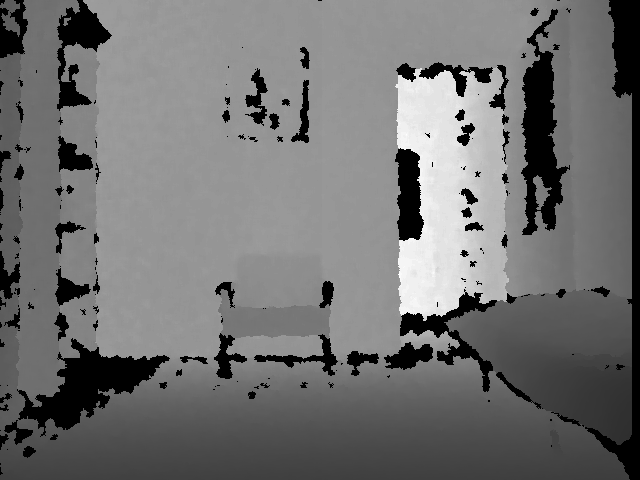}
  \end{subfloat}
    \begin{subfloat}[]
    \centering
    \includegraphics[scale=0.1]{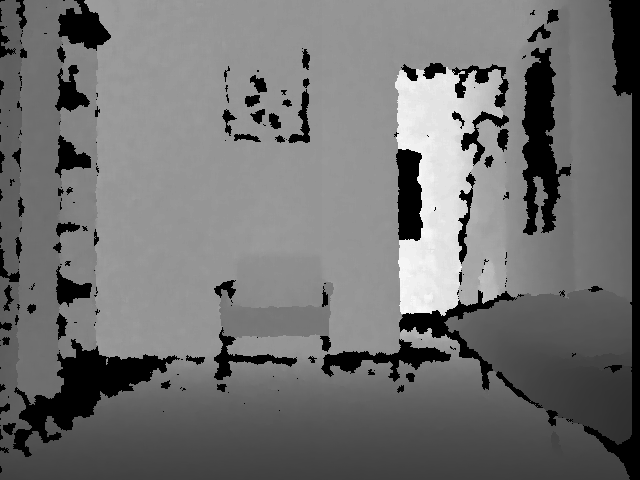}
  \end{subfloat}
  \begin{subfloat}[]
    \centering
    \includegraphics[scale=0.1]{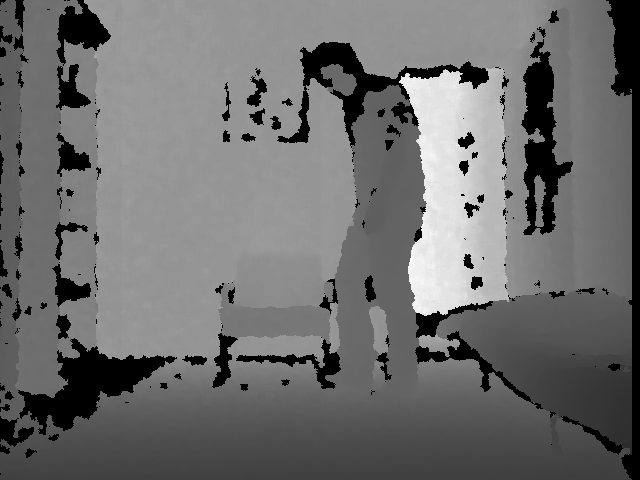}
  \end{subfloat}
  \begin{subfloat}[]
    \centering
    \includegraphics[scale=0.1]{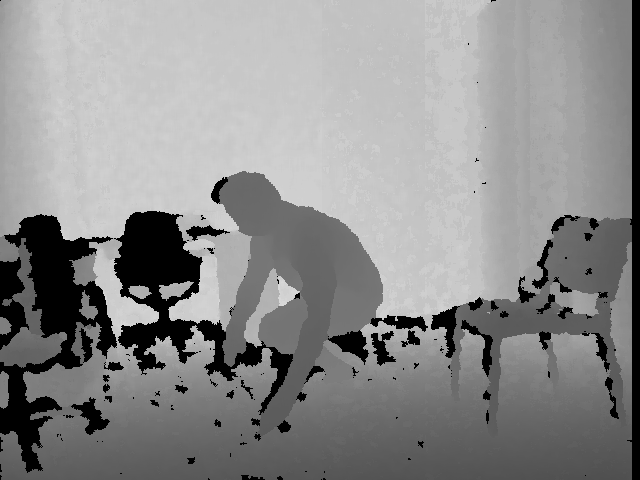}
  \end{subfloat}
  \caption{UR Dataset - Original Depth frames with holes (a) Empty Scene (b) Person entering the Scene, (c) Person in the Scene, (d) Fall.}
\label{fig:URHoles}
  \vspace{5mm}  
    \begin{subfloat}[]
    \centering
    \includegraphics[scale=0.1]{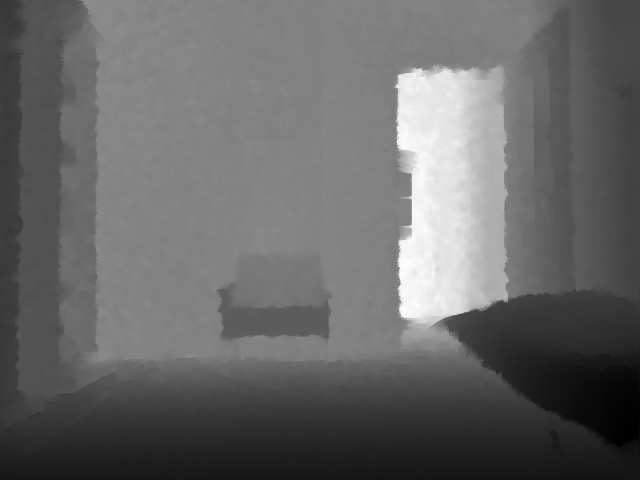}
  \end{subfloat}
    \begin{subfloat}[]
    \centering
    \includegraphics[scale=0.1]{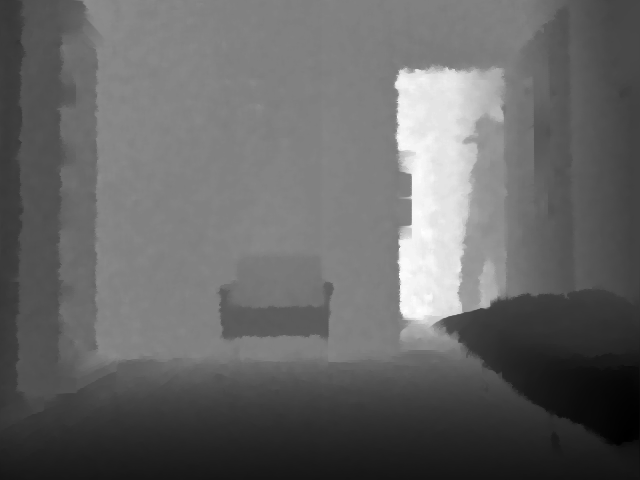}
  \end{subfloat}
    \begin{subfloat}[]
    \centering
    \includegraphics[scale=0.1]{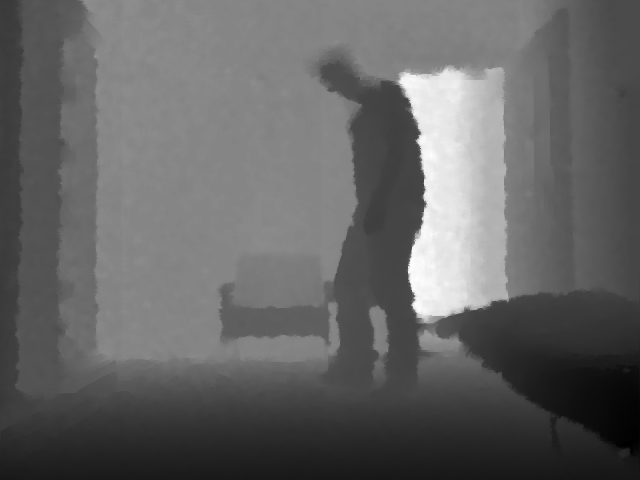}
  \end{subfloat}
  \begin{subfloat}[]
    \centering
    \includegraphics[scale=0.1]{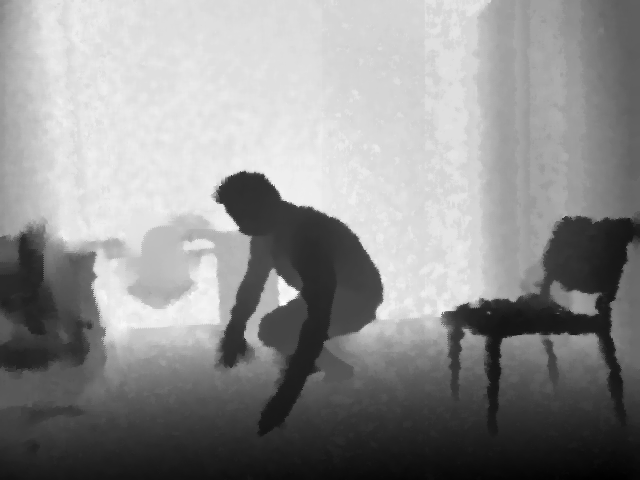}
  \end{subfloat}
    \caption{UR Dataset - Depth frames after holes filling (a) Empty Scene (b) Person entering the Scene, (c) Person in the Scene, (d) Fall.}
    \label{fig:URHolesFilled}
\end{figure}

\item \textit{SDU Fall Dataset}:
    In the SDU Fall dataset \cite{SDU}, ten young men and women did six types of activities $30$ times, resulting in $1800$ video clips. 
    The data shared with us contained $1197$ videos, out of which $997$ had normal ADL and $200$ had falls. The activities included falling, bending, squatting, sitting, lying, and walking. These videos were captured using a Kinect camera (thus hiding person's identity) at $30$fps, with video frame size of $320\times240$ and stored in AVI format. The SDU fall dataset also had holes similar to the UR dataset. However, the information on distance of depth frames is not provided with this dataset; therefore, we used an inpainting approach \cite{NS} to fill these holes, we call that data as SDU-filled. After applying the sliding window ($T=8$), $163,573$ windows of contiguous frames were obtained. A sample of normal ADL and falls from SDU and SDU-filled is shown in Figures \ref{fig:SDUHoles} and \ref{fig:SDUHolesFilled}.
    
    
\begin{figure}[!ht]
\centering
\captionsetup[subfigure]{width=80pt}%
  \begin{subfloat}[]
    \centering
    \includegraphics[scale=0.1]{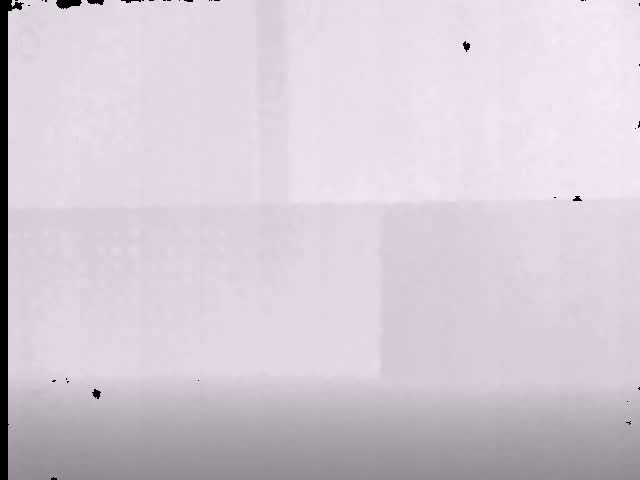}
  \end{subfloat}
 \captionsetup[subfigure]{width=80pt}%
  \begin{subfloat}[]
    \centering
    \includegraphics[scale=0.1]{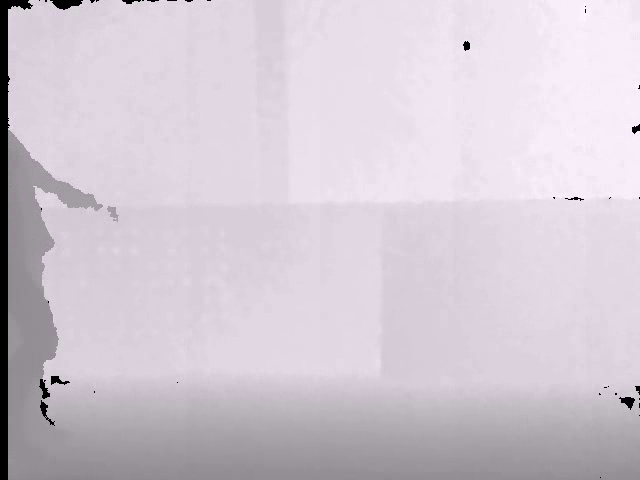}
  \end{subfloat}
  \begin{subfloat}[]
    \centering
    \includegraphics[scale=0.1]{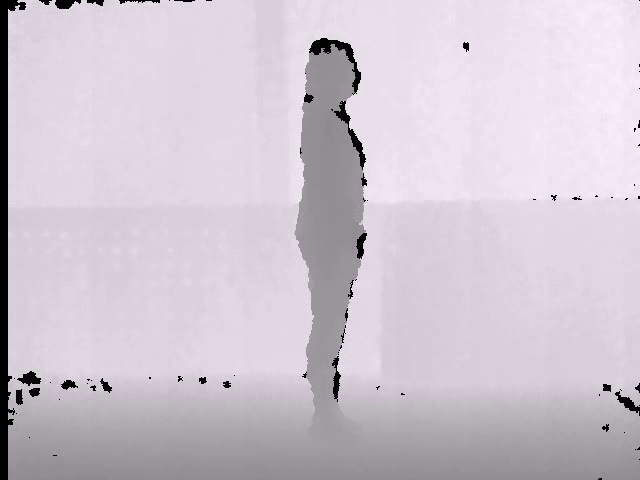}
  \end{subfloat}
   \begin{subfloat}[]
    \centering
    \includegraphics[scale=0.1]{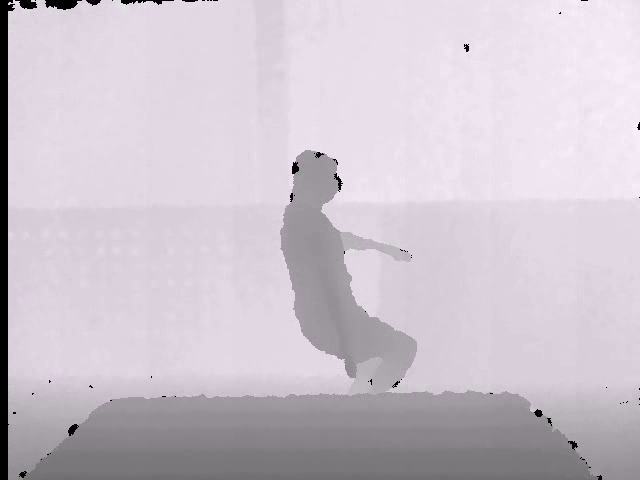}
  \end{subfloat}
  \caption{SDU Dataset - Original Depth frames with holes (a) Empty Scene (b) Person entering the Scene, (c) Person in the Scene, (d) Fall.}
  \label{fig:SDUHoles}
\vspace{5mm}  
\begin{subfloat}[]
    \centering
    \includegraphics[scale=0.1]{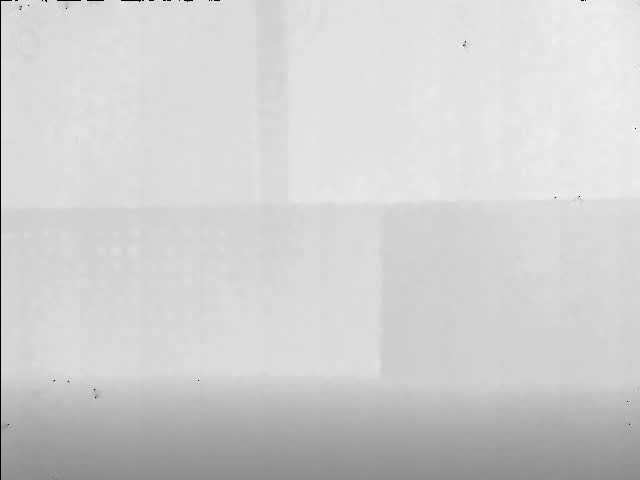}
  \end{subfloat}
 \captionsetup[subfigure]{width=80pt}%
  \begin{subfloat}[]
    \centering
    \includegraphics[scale=0.1]{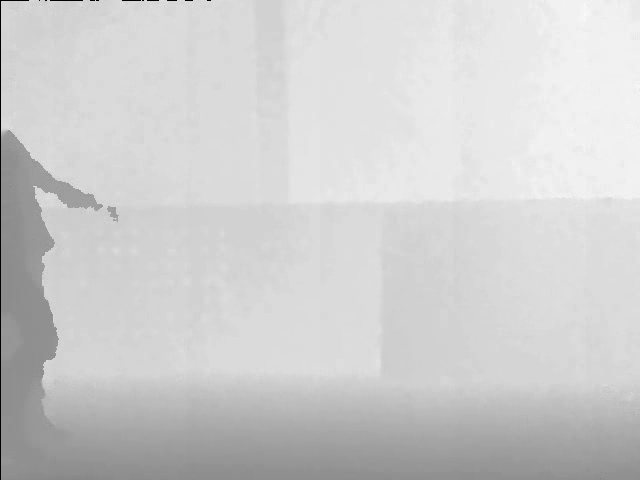}
  \end{subfloat}
  \begin{subfloat}[]
    \centering
    \includegraphics[scale=0.1]{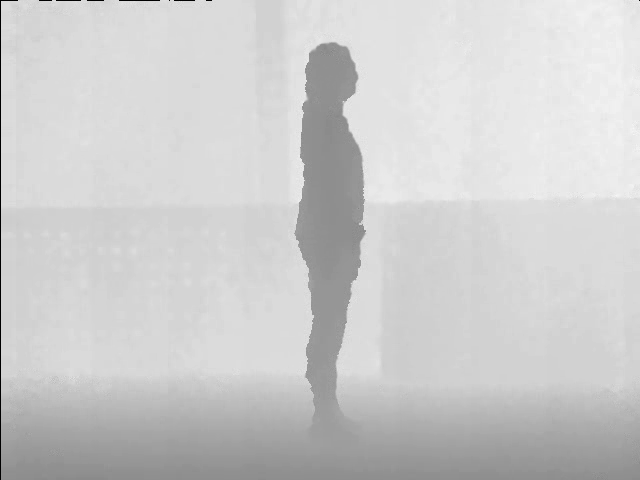}
  \end{subfloat}
  \begin{subfloat}[]
    \centering
    \includegraphics[scale=0.1]{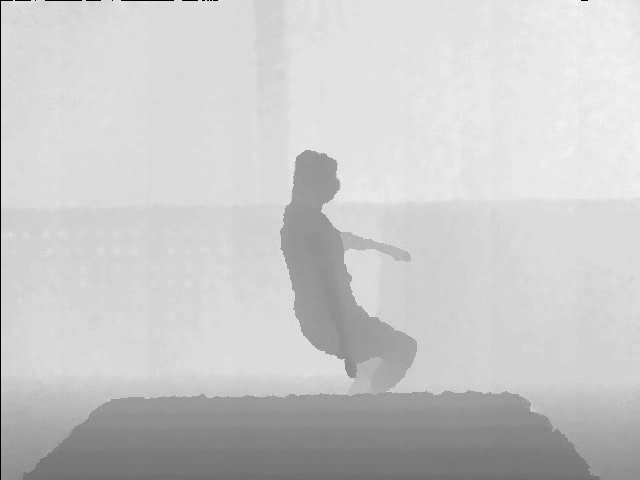}
  \end{subfloat}
  \caption{SDU Dataset - Depth frames after hole filling, (a) Empty Scene (b) Person entering the Scene, (c) Person in the Scene, (d) Fall.}
  \label{fig:SDUHolesFilled}
\end{figure}

\end{enumerate}

In all the datasets, there are empty frames with no person,  with person entering from left, right, front far end or a full person in the scene. All the frames in all the datasets are resized to $64\times64$, normalized by dividing the pixel value by $255$ to keep them in the range $[0,1]$, then subtracting the per-frame means from each frame, which keeps the pixel values in the range $[-1,1]$. The different adversarial trained methods are trained on only the normal ADL frames or their sequences. For testing, videos are presented to the trained network containing both normal ADL and unseen fall frames (or their sequences), which were manually annotated as ground truth. 
Since a fall is a short event, it can only take few frames for a fall event from start to end. In our datasets, the maximum number of frames for a fall to occur was $13$. Since we wanted to keep the number of frames to be a power of $2$; therefore, we choose $T=8$ as higher values of $T$ would not be possible. Smaller values of T$=4$ resulted in more false alarms and their results are not shown in the paper.
In our implementation of the spatio-temporal adversarial learning, we use SGD optimizer with learning rate equals to $0.0002$ for the 3DCNN discriminator, and adadelta optimizer for the 3DCAE. 
We trained our model on various values of $\lambda$. Larger values of $\lambda$ lead to mode collapse problem. Therefore, we choose $\lambda=1$ that gave the best results. We train all the adversarial methods for a maximum of $500$ epochs.

\subsection{Results}
\paragraph{Frame Level Anomaly}: Table \ref{tab:framelevel} shows the Area Under the Curve (AUC) values after applying frame level anomaly scoring method on DAE-AN, CAE-AN and the proposed spatio-temporal adversarial network (on $C_\mu$ and $C_\sigma$ anomaly score).  The best AUC values are shown in gray color cells. We observe that the proposed method performs better than DAE-AN and CAE-AN on all the datasets, except SDU-filled with DAE-AN.
The SDU dataset videos contains simple and organic activities, falls always happened from standing, besides having no furniture or background objects in the scene. We hypothesize that due to these reasons the DAE-AN and CAE-AN may be able to learn global and spatial features that may be able to detect falls comparable to the spatio-temporal network. However, the activities in the Thermal and UR datasets were complex, falls happened in various poses (e.g. falling from chair, falling from sitting and falling from standing), and the scene involved different objects in the background (e.g. bed, chair). In the Thermal dataset, due to a person entering the scene, the pixel intensity would change values due to change in the heat in the environment. The proposed spatio-temporal adversarial learning method worked well under these diverse condition to detect unseen falls.
We also observe that all the fall detection methods performed worse on original UR and SDU datasets than their holes filled versions. This clearly shows that videos with holes are detrimental to learn normal ADL and identify unseen falls.
We further observe that AUC results of the proposed approach are slightly better with $C_\sigma$ than $C_\mu$ for all the datasets. 

\begin{table}[!ht]
\centering
\resizebox{\columnwidth}{!}{%
\begin{tabular}{|c||c||c|c||c|c|}
\hline
\multirow{2}{*}{Models}  & \multicolumn{5}{|c|}{Datasets} \\ \cline{2-6}
  & Thermal & UR & UR-Filled & SDU & SDU-Filled \\ \hline\hline
DAE-AN & 0.62 &  0.46 & 0.65 & 0.68 &\cellcolor{gray!50}\textbf{0.91}\\ \hline
\hline
CAE-AN & 0.62 &  0.36 & 0.78 & 0.62 & 0.89\\ \hline
\hline
$C_{\mu}$ & \cellcolor{gray!50}\textbf{0.95} &  0.47 & 0.88 & 0.69 & 0.90\\ \hline
$C_{\sigma}$ & \cellcolor{gray!50}\textbf{0.95} &  0.74 & \cellcolor{gray!50}\textbf{0.91} & 0.69 & \cellcolor{gray!50}\textbf{0.91}\\ \hline

\end{tabular}
}
\caption{AUC values for different adversarial networks for each dataset (using frame based anomaly scoring).}
\label{tab:framelevel}
\end{table}

\begin{figure}[!ht]
\vspace{-3mm}
    \centering
    \includegraphics[scale=0.5]{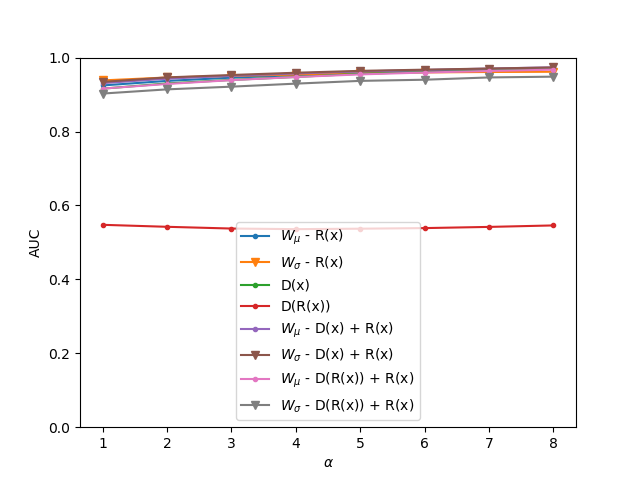}
    \vspace{-5mm}
    \caption{Variation of AUC w.r.t $\alpha$ on Thermal dataset for different models (using window based anomaly scoring).}
    \label{fig:thermalW}
\end{figure}

\begin{figure}[!ht]
\vspace{-7mm}
    \centering
    \includegraphics[scale=0.5]{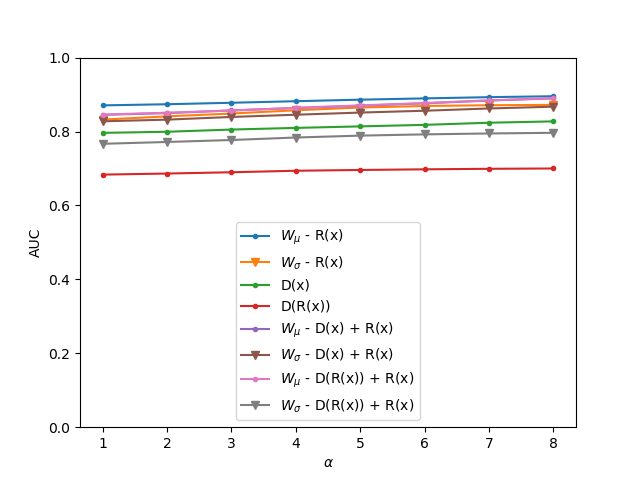}
    \vspace{-5mm}
    \caption{Variation of AUC w.r.t $\alpha$ on UR-Filled dataset for different models (using window based anomaly scoring).}
    \label{fig:urfilledW}
\end{figure}

\begin{figure}[!ht]
\vspace{-7mm}
    \centering
    \includegraphics[scale=0.5]{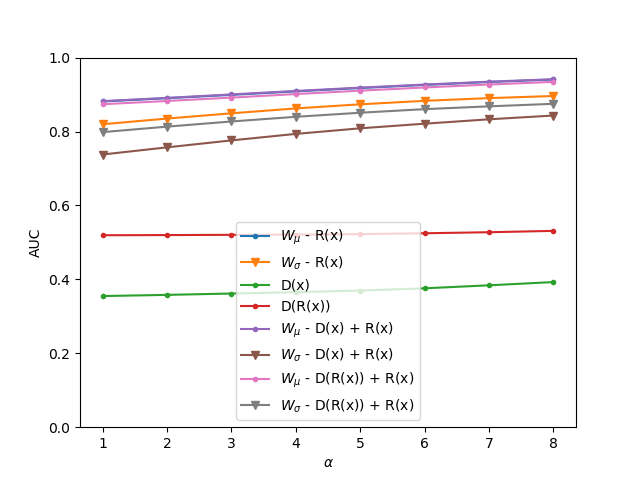}
    \vspace{-5mm}
    \caption{Variation of AUC w.r.t $\alpha$ on SDU-Filled dataset for different models (using window based anomaly scoring).}
    \label{fig:sdufilledW}
\end{figure}

\paragraph{Window Level Anomaly}:
Figures \ref{fig:thermalW}, \ref{fig:urfilledW} and \ref{fig:sdufilledW} show the AUC values of the spatio-temporal adversarial learning on detecting unseen falls on Thermal, UR-Filled and SDU-Filled datasets using window level anomaly scores w.r.t. different choices of $\alpha$ from $1$ to $8$ (which is the maximum size of the window) . The results on UR and SDU with holes were consistently worse and are not shown. We observe that for different anomaly scores for each of the datasets, the AUC initially increases with an increase in the number of fall frames in a window (i.e. $\alpha$) and then stabilizes for higher values of $\alpha$. This is related to the fact that if a window is decided as a `fall' based on very few fall frames, it would lead to many false alarms, resulting in lower AUC. It can be clearly seen that the anomaly score $D(R(x))$ performs worst in all the datasets. Furthermore, in UR-Filled datasets, the two other worse performing scores are $W_\sigma - D(R(x))+R(x)$ and $D(x)$, and in SDU-Filled are $D(x)$ and $W_\sigma - D(x)+ R(x)$. Other anomaly scores perform equivalent to each other. This experiment suggests that unseen falls can be detected with high AUC using window level anomaly scoring. However, the scores obtained at the discriminator or when combined with reconstruction error may not be a good candidate for detecting unseen falls. 

It is to be noted that the scores of window level anomaly scoring are not directly comparable with frame level scoring method. In the frame level method, the anomaly score is calculated for every frame (occurring at different windows). Whereas in the window level method, we designate the class of the whole window instead of deciding the class of every frame across windows. Another factor in window level anomaly is the number of fall frames present in a window ($\alpha$), s.t. the ground truth label of the entire window is a fall. This parameter is varied and results are shown in Figures \ref{fig:thermalW}, \ref{fig:urfilledW} and \ref{fig:sdufilledW}. Therefore, these two types of anomaly scoring methods are not directly comparable and their separate results are discussed in the paper.

The proposed framework may detect other abnormal activities as falls that significantly deviates from normal ADL, such as syncope, tripping, or presence of new objects or people in the scene. However, on the datasets we tested, those variation were not present.

\section{Conclusions and Future Work}
\label{sec:conclusions}
This paper deals with identifying unseen falls in videos using a new spatio-temporal adversarial learning framework. The videos used in this paper are privacy preserving, such as thermal and depth cameras that can partially or fully obfuscate facial features of a person. This further ascertains the idea that for fall detection problem, only spatial and temporal information contained in the video is needed and not the identity revealing information (e.g. face of the person). We present a learning strategy to train the adversarial framework using spatio-temporal autoencoder and a spatio-temporal discriminator. 
The results on three public datasets suggest high performance in comparison to two other spatial adversarial baselines. Encouraged by the results presented in this paper, we are currently collecting a new dataset on fall detection using multiple types of vision sensing modalities, such as thermal cameras, depth cameras, an IP camera and a RGB camera. These ceiling mounted cameras represent a more realistic scenario of using them in a home-setting. This unique dataset will be made public and will help us comparing different sensing modalities for the problem of fall detection. Furthermore, in future, we will use spatio-temporal residual / U-net networks with attention in an adversarial framework to detect unseen falls and other health related abnormal behaviours.

\bibliographystyle{spmpsci}
\bibliography{references}
\end{document}